# Rice Classification Using Spatio-Spectral Deep Convolutional Neural Network

Itthi Chatnuntawech[1], Kittipong Tantisantisom[1], Paisan Khanchaitit[1], Thitikorn Boonkoom[1], Berkin Bilgic[2,3], Ekapol Chuangsuwanich[4]

[1]National Nanotechnology Center, National Science and Technology Development Agency, Pathum Thani, Thailand
[2]Athinoula A. Martinos Center for Biomedical Imaging, Charlestown, MA, USA
[3]Harvard Medical School, Boston, MA, USA
[4]Computer Engineering Department, Chulalongkorn University, Bangkok, Thailand

## Abstract

Rice has been one of the staple foods that contribute significantly to human food supplies. Numerous rice varieties have been cultivated, imported, and exported worldwide. Different rice varieties could be mixed during rice production and trading. Rice impurities could damage the trust between rice importers and exporters, calling for the need to develop a rice variety inspection system. In this work, we develop a non-destructive rice variety classification system that benefits from the synergy between hyperspectral imaging and deep convolutional neural network (CNN). The proposed method uses a hyperspectral imaging system to simultaneously acquire complementary spatial and spectral information of rice seeds. The rice varieties are then determined from the acquired spatio-spectral data using a deep CNN. As opposed to several existing rice variety classification methods that require hand-engineered features, the proposed method automatically extracts spatio-spectral features from the raw sensor data. As demonstrated using two types of rice datasets, the proposed method achieved up to 11.9% absolute improvement in the mean classification accuracy, compared to the commonly used classification methods based on support vector machines.

## 1. Introduction

Rice has been one of the most widely consumed foods for a large part of human population. Numerous different rice varieties are imported and exported worldwide, making it the backbone of many countries' economy. Rice seeds of different varieties can be accidentally or intentionally mixed during any of the steps in a rice production pipeline, introducing impurities. These impurities could damage the trust between rice importers and exporters, calling for the need to develop a reliable rice variety inspection system.

Several methods based on biological or chemical techniques such as genetic markers have been proposed to determine rice varieties (Steele et al. 2008; Cirillo et al. 2009; Chuang et al. 2011). Although these methods are very accurate, they are destructive techniques that are costly and time-consuming, making them unsuitable for mass inspection. Being limited to batch sampling, these techniques cannot truly assess the purity of the rice seeds under inspection. To circumvent these limitations, non-destructive rice inspection systems that use a combination of optical imaging and multi-variate data analysis techniques have been proposed (Liu et al. 2005, 2010; Guzman and Peralta 2008; OuYang et al. 2010; Mousavi Rad et al. 2012; Kong et al. 2013; Pazoki et al. 2014; Hong et al. 2015; Sun et al. 2015; Wang et al. 2015; Vu et al. 2016; Kuo et al. 2016; Huang and Chien 2017). For these techniques, the data from rice seeds are acquired using an optical imaging system. Some information that characterizes the acquired data (commonly called features) is then extracted as input for a data classification algorithm.

There are two types of features that are frequently extracted for these methods: spatial features and spectral features. Common spatial features that describe the visual appearance of a rice seed include shape, morphological, and textural features. Several research groups have demonstrated promising classification accuracies of the rice varieties that are visually distinguishable, using only spatial features acquired with a digital camera (Liu et al. 2005; Guzman and Peralta 2008; OuYang et al. 2010; Mousavi Rad et al. 2012; Pazoki et al. 2014; Hong et al. 2015; Huang and Chien 2017) and with some additional equipment such as a microscope (Kuo et al. 2016).

While a typical digital camera provides only partial information in the visible range, other types of equipment such as a hyperspectral imaging camera can be used to acquire information from a wider range of the electromagnetic spectrum. A typical hyperspectral imaging camera can provide both spatial and spectral information in a specific portion of the electromagnetic spectrum. For example, a spectrum that contains the information on chemical properties at each spatial location can be acquired using a near-infrared hyperspectral imaging camera. It has been demonstrated that rice panicles classification (Liu et al. 2010) and rice variety classification (Kong et al. 2013) can be performed successfully by using only the spectral information acquired with a hyperspectral imaging system. Recently, rice classification systems that use both spatial and spectral features have been proposed (Sun et al. 2015; Wang et al. 2015; Vu et al. 2016). These systems use a hyperspectral imaging camera to acquire data with both spatial and spectral information. They have demonstrated that using both spatial and spectral features resulted in higher classification accuracies, compared to the case when only spatial or spectral features were used.

Although the existing works have shown promising results, they rely on hand-engineered features that require extensive domain specific knowledge. Inspired by the desire to bypass a predefined feature extraction step, deep learning, a subfield of machine learning that stems from artificial neural networks, has emerged as an alternative to conventional classification methods. Having an ability to represent data with multiple levels of abstractions, deep learning has been demonstrated to be comparable to, or surpass existing methods, achieving state-of-the-art performances in a wide range of applications such as image recognition (Krizhevsky et al. 2012; He et al. 2016; Huang et al. 2017), speech recognition (Mikolov et al. 2011; Dahl et al. 2012; Hinton et al. 2012; Graves et al. 2013), medical image processing and analysis (Ronneberger et al. 2015; Esteva et al. 2017; Poplin et al. 2018), hyperspectral imaging (Chen et al. 2014; Hu et al. 2015), and rice variety identification (Lin et al. 2018; Patel and Joshi 2018; Qiu et al. 2018).

The purpose of this work was to develop a non-destructive technique to improve the accuracy of rice variety classification. The proposed technique uses a hyperspectral imaging system to simultaneously acquire spatial and spectral information of rice seeds. The rice varieties are then determined from the acquired spatio-spectral data using a spatio-spectral classification method that is developed based on a deep convolutional neural network (CNN). While the existing deep learning-based rice classification methods only process information in either the spatial domain (Lin et al. 2018; Patel and Joshi 2018) or the spectral domain (Qiu et al. 2018), the proposed method simultaneously exploits the information from both the spatial and spectral domains. Compared to the existing rice variety classification methods that require hand-engineered feature extractions (Liu et al. 2005, 2010; Guzman and Peralta 2008; OuYang et al. 2010; Mousavi Rad et al. 2012; Kong et al. 2013; Pazoki et al. 2014; Hong et al. 2015; Sun et al. 2015; Wang et al. 2015; Vu et al. 2016; Kuo et al. 2016; Huang and Chien 2017), the proposed method automatically extracts features from the data. Furthermore, while the existing traditional methods involve multiple subsequent data processing steps and hence suffer from error propagation between the subsequent steps, the proposed classification method is an end-to-end method that does not encounter such a problem.

## 2. Materials and Methods

The proposed rice variety classification systems consist of two parts: data acquisition, and data processing and classification. In this work, as shown in Figure 1, we used a near-infrared hyperspectral imaging system to acquire data because it is a rapid data acquisition method that provides both spatial and spectral information without damaging the samples under inspection. The acquired spatio-spectral data, which are commonly called datacubes, had three dimensions: two spatial dimensions and one spectral dimension. While the spatial dimensions provided information on the visual appearances of the rice seeds, the spectral dimension provided complementary information on their chemical properties. From the acquired datacubes, the rice varieties were then determined using a data processing and classification method.

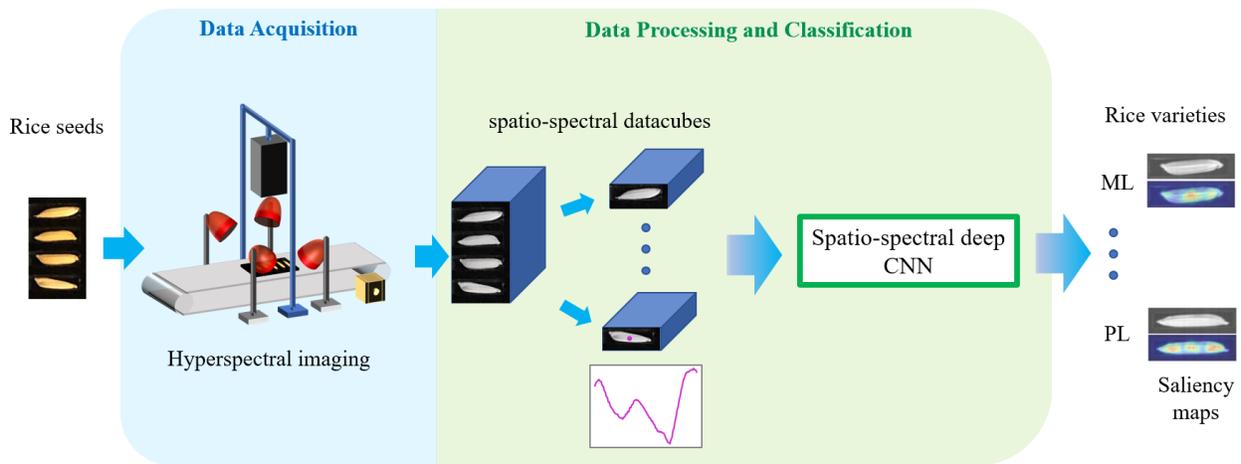

Figure 1. The proposed rice variety classification system based on a hyperspectral imaging system and a spatio-spectral deep convolutional neural network.

## 2.1 Data Acquisition

### 2.1.1 Hyperspectral Imaging System

A line-scan hyperspectral imaging system was constructed as depicted in Figure 2 in a dark room. The system consisted of the following components: (1) a near-infrared hyperspectral imaging camera that covers the 900 - 1700 nm spectral range with a spectral resolution of 5.49 nm (Pika NIR, Resonon Inc., USA), (2) four 50-watt halogen lamps to illuminate the region of interest (GU5.3, Philips, Thailand), (3) a conveyor belt (Oriental Motor Co., Ltd., Japan), (4) a 90W, US-52 speed controller for the conveyor belt (AP Electric, Thailand), and (5) a computer to collect and process the acquired data.

### 2.1.2 Datacube Acquisition

Prior to data acquisition, the halogen lamps were preheated with applied bias of 6.8 volts for 30 minutes to allow them to reach a stable operating condition. The datacubes of rice seeds were acquired using the line-scan hyperspectral imaging camera. Only one spatial dimension (one line) and the spectral dimension of the rice seeds that were located right under the hyperspectral imaging camera were acquired at a time. The working distance (i.e., the distance between the lens and samples) was set to 28 cm. In order to collect the data along the second spatial dimension, the conveyor belt was used to move the rice seeds to different locations over time. The speed controller was used to control the conveyor belt's speed to maintain a realistic aspect ratio of the acquired data. The data from the hyperspectral camera were then sent to the computer for further data processing. The datacubes of all the rice seeds were acquired in the same geometric orientation as shown in Figure 2.

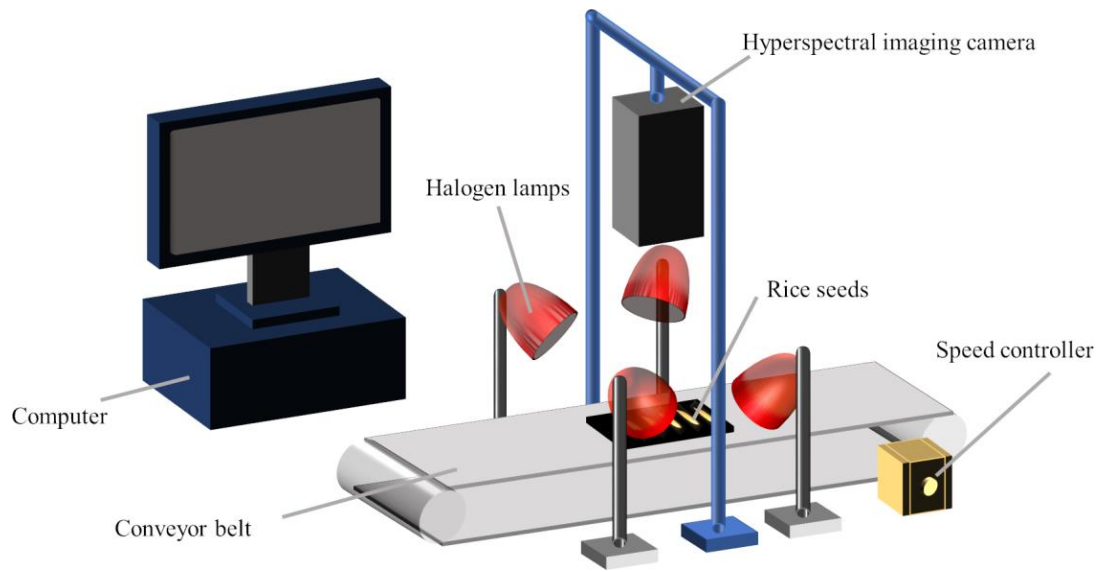

Figure 2. The line-scan hyperspectral imaging system used for data acquisition.

## 2.2 Data Processing and Classification

**Spatio-Spectral Deep Convolutional Neural Network**

A deep CNN is a class of machine learning model that was invented with an inspiration from visual neurosciences (Hubel and Wiesel 1962). A deep CNN consists of a series of processing layers. As the data enter the network, they go through a cascade of linear and nonlinear mathematical operations of each processing layer. The processed data from one layer are used as inputs for the next layer. Each layer transforms the representation of its inputs into another representation that is more complex and abstract. Consequently, a deep CNN learns how to represent data with multiple levels of abstractions by itself, as opposed to conventional classification methods, which rely on predefined feature extractions that have been developed from human past experiences. Please refer to Refs. (Lecun et al. 2015; Goodfellow et al. 2016), for example, for fundamentals of and recent advances in deep learning.

In this work, we developed a data classification method based on a deep CNN with hundreds of processing layers. We modified and extended a residual network (ResNet) with bottleneck building blocks (He et al. 2016) to enable spatio-spectral data classification. The proposed method (abbreviated as ResNet-B) takes advantages of residual connections together with bottleneck building blocks (He et al. 2016), batch normalization (Ioffe and Szegedy 2015), and early stopping to aid the training process of such a deep network. Residual connections help improve the information flow throughout the network, which results in easier training processes of neural networks, especially for the deep ones. Bottleneck building blocks help increase representation power by allowing more layers to be added to the network without increasing the number of parameters in the network (He et al. 2016). Batch normalization helps mitigate the covariance shift problem (Shimodaira 2000), which allows us to use high values of the learning rate, accelerating the training process. Early-stopping is used to prevent the network from overfitting the data, and hence potentially increases the classification accuracy of unseen samples.

Most existing rice variety classification methods involve a separate feature extraction process that consists of multiple sequential data processing steps. Each of the sequential steps is usually associated with some adjustable parameters, and requires good parameter tuning to obtain acceptable performance. If suboptimal parameters are chosen for one of the steps, it could adversely affect the performances of all the subsequent steps, including the classification step. To prevent such error propagation, we used a deep CNN to simultaneously perform feature extractions and rice variety classification. The entire datacube is used as features, removing the need for multi-step feature extractions.

Another key advantage of the proposed method is that it can automatically discover the features needed for classification. Without relying on hand-engineered features that require extensive domain specific knowledge and good feature engineering skill, the proposed method can potentially discover data presentations that truly matter for classification automatically. Moreover, while recent rice classification methods treat spatial feature extraction and spectral feature extraction as two separate processes, and construct "spatio-spectral" features by simply concatenating the extracted spatial and spectral features (Sun et al. 2015; Wang et al. 2015; Vu et al. 2016), the proposed method could create genuine spatio-spectral features because it performs feature extraction directly on the datacube, which provides direct access to both spatial and spectral information.

## 2.3 Main Experiments

Using two types of rice datasets, we compared the performance of the proposed method (ResNet-B) to three different variants of the method most commonly used for rice classification: support vector machine (SVM) with spatial features, SVM with spectral features, and SVM with both spatial and spectral features. Please refer to the Appendix for more details regarding the SVM-based methods.

### 2.3.1 Datasets

Two rice datasets, consisting of processed rice and paddy rice datasets, were acquired from the hyperspectral imaging system and used to assess the performances of the classification methods. The processed rice dataset served as a proof-of-concept since the varieties were easily distinguishable by the visual appearances and/or spectral profiles of the rice seeds. The paddy rice dataset represented a real-world scenario in Thailand, where the rice varieties in this dataset have been frequently mistaken for the others because they have very similar visual appearances and spectral profiles, posing a serious challenge to rice inspection systems. In this case, without an accurate inspection system, relatively low quality rice varieties could be mistaken for high quality rice varieties and sold at higher-than-normal prices, damaging the trust in rice import and export industries. Figure 3 shows example rice seeds from each rice variety in the datasets along with the mean spectra.

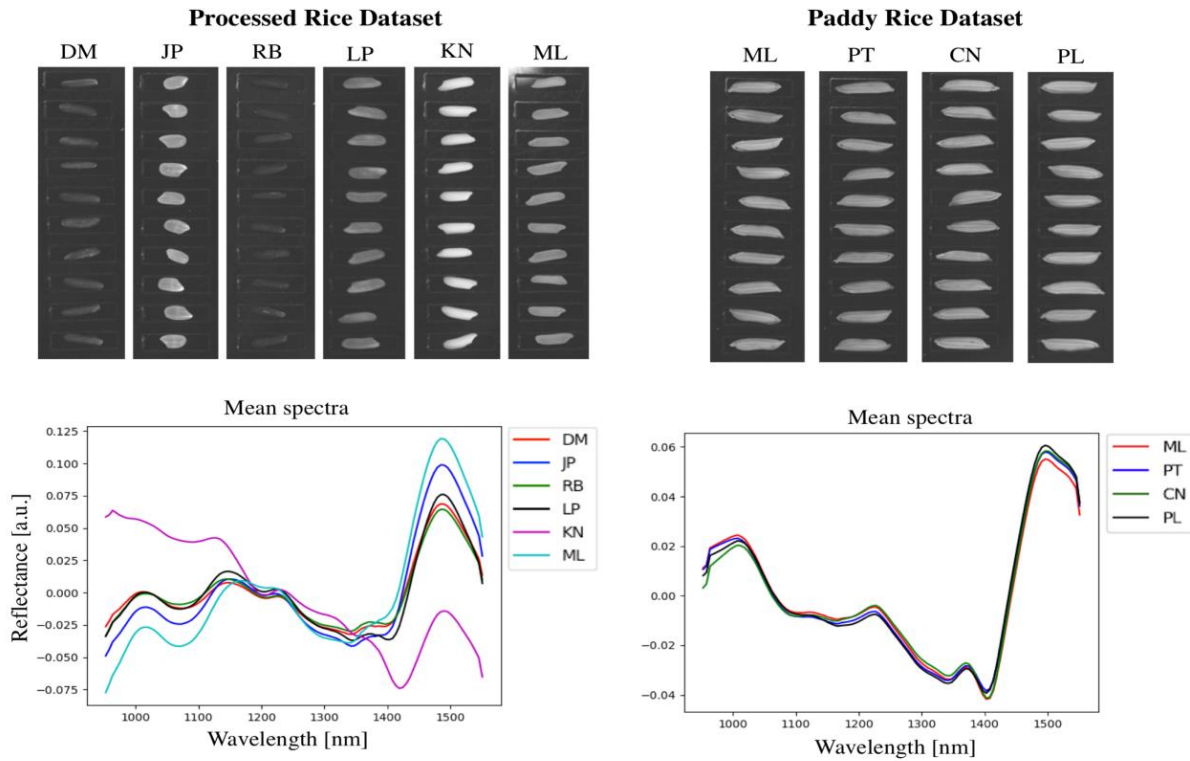

Figure 3. Examples of the rice seeds from each rice variety in the processed (left) and paddy (right) datasets along with the mean reflectance spectra (after being demeaned and filtered).

### 2.3.1.1 Processed Rice

In this work, processed rice referred to both brown rice (paddy rice with outer hull removed) and white rice (paddy rice with outer hull and almost all bran and germ removed). Six varieties of processed rice were obtained from a seed production company in Thailand: Daeng Mun Pooh (DM), Uruchimai Japonica (JP), Riceberry (RB), Leuang Patew (LP), Kiaw Ngu (KN), and Khao Dawk Mali 105 (i.e., Thai jasmine rice, ML). Using the hyperspectral imaging system, 232 datacubes were acquired for each rice variety (one datacube per rice seed), resulting in a total of 1,392 datacubes. Each acquired datacube was a 50 x 170 x 110 tensor. Specifically, each datacube consisted of images of size 50 x 170 pixels with each pixel containing 110 spectral bands, ranging from 950 nm to 1550 nm.

### 2.3.1.2 Paddy Rice

Four Thai rice varieties were obtained from the Thai Rice Department of the Ministry of Agriculture and Cooperatives: Khao Dawk Mali 105 (ML), Pathum Thani 1 (PT), Chai Nat 1 (CN), and Phitsanulok 2 (PL). Compared to the first dataset, these varieties had much more similar visual appearances and spectral profiles. Using the same hyperspectral imaging system, 414 datacubes were acquired for each rice variety, resulting in a total of 1,656 datacubes. Each acquired datacube was a 50 x 170 x 110 tensor.

### 2.3.2 Implementation Details

The acquired datacubes of the rice seeds were randomly split into training and test data: 85% for training and 15% for testing. The training data were used to optimize model parameters and train a classifier. The test data were used to assess the performance of the trained classifier. We performed the training-test data splitting multiple times, and then reported the mean and standard deviation (over the repetitions) of the classification accuracies on the test data. All the data processing and classification methods were implemented in Python, and run on a workstation with 48 Intel Xeon Gold 5118 processors, 128 GB of memory, and a 12 GB NVIDIA Tesla P100 graphics card. OpenCV (Bradski and Kaehler 2008), Scikit-learn (Pedregosa et al. 2011), and Scikit-image (Van der Walt et al. 2014) were mainly used for the SVM-based methods. Keras (Chollet 2015; Kotikalapudi 2017) with the Tensorflow (Abadi et al. 2016) backend was used for the proposed method. Example data and implementations can be downloaded from https://github.com/ichatnun/spatio-spectral-resnet-bottlenecks-rice-classification.

For the proposed method, prior to data classification, each datacube was normalized by its maximum value. Then, a deep CNN was used to determine the rice variety of each rice seed. The spectral dimension of each datacube was treated as the channel dimension of the input to the CNN. Specifically, the input datacubes (50 x 170 x 110 tensors) can be thought of as two-dimensional images of size 50 x 170 with 110 channels (as opposed to only three channels, RGB, in a typical optical imaging camera).

The original training data (85% of the acquired datacubes) were further split into the validation and new training data with proportions of 20% and 65% of the acquired datacubes, respectively. The validation data were used to select some of the hyperparameters: initial learning

rate = 0.005, batch size = 4, activation function = Swish (Ramachandran et al. 2017) (also called ESP in (Milletarí et al. 2018)), and number of nodes in the classification layer = 512 and 1024 for the processed rice and paddy rice datasets, respectively. Following the practice in the original ResNet paper (He et al. 2016), we did not use dropout. The exact network configurations of ResNet-B is shown in Figure 4. After the hyperparameters had been selected, the internal parameters of the networks (e.g., weights and biases) were optimized by minimizing the cross-entropy cost function on both the new training data and validation data for 400 epochs using the Adam optimizer (Kingma and Ba 2014) with the following parameters defined in the Keras documentation: learning rate = initial learning rate divided by the batch size, $\beta_1$=0.9, $\beta_1$=0.999, $\epsilon$=$10^{-8}$, and learning rate decay=0.01.

In this work, data augmentation techniques (Yaeger et al. 1996; Simard et al. 2003; Cireşan et al. 2011; Krizhevsky et al. 2012) were employed to increase the amount of training data without the need to acquire more data through the data acquisition process. Specifically, more datacubes were generated by retrospectively modifying the existing datacubes with some mathematical operations consisting of shifting and flipping the datacubes. Both horizontal and vertical flippings were used. Moreover, the datacubes were randomly shifted by different amounts in the spatial domain. The maximum shifting amount was set to 4% of the number of pixels in each spatial dimension. Combining these operations, we increased the amount of training datacubes by more than 10-fold. Figure 5 shows a simplified flowchart of the proposed method.

## 2.4 Deep Learning Experiments

In addition to the main experiments where we compared ResNet-B to the SVM-based methods, a deep learning experiment was conducted to demonstrate the feasibility of several other spatio-spectral deep convolutional neural networks for the rice variety classification task. In particular, we compared the performances of four spatio-spectral deep CNNs on the paddy rice dataset: VGGNet (Simonyan and Zisserman 2015), ResNet without bottleneck building blocks, ResNet with bottleneck building blocks (ResNet-B) (He et al. 2016), and an ensemble of VGGNet, ResNet, and ResNet-B.

The network architectures of VGGNet, ResNet, and ResNet-B are shown in Figure 4 and summarized in Table 1. While the number of parameters of VGGNet, ResNet, and ResNet-B were kept comparable for fair comparisons, ResNet-B has many more convolutional layers. Apart from the slightly different network architectures between VGGNet, ResNet, and ResNet-B, the remaining hyperparameters were kept the same as those described in the Main Experiments subsection. We observed that it was very difficult to train the networks and achieved reasonable results without batch normalization (results not shown). Consequently, we employed batch normalization for all the methods in this experiment to ease the training processes. An ensemble of VGGNet, ResNet, and ResNet-B was also implemented by combining the predictions of the three networks. Specifically, the mean of the outputs of the softmax layer from the three networks was treated as the prediction of the ensemble.

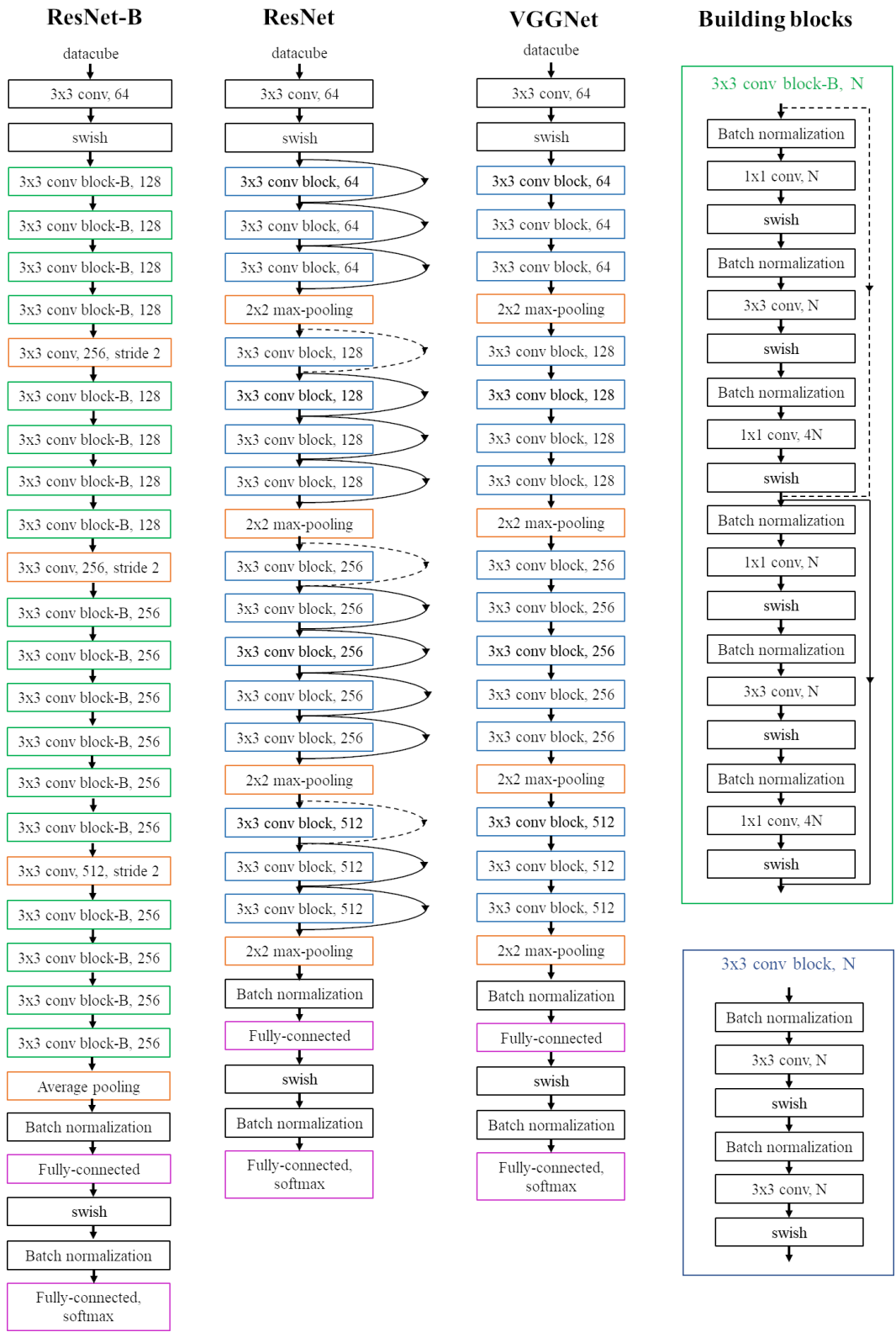

Figure 4. Network architectures of ResNet-B, ResNet, and VGGNet with their corresponding building blocks. The dashed lines represent residual connections that possibly contain a 1x1 convolutional layer to avoid the dimension mismatch problem in the channel axis. All the convolutional layers had a stride of one.

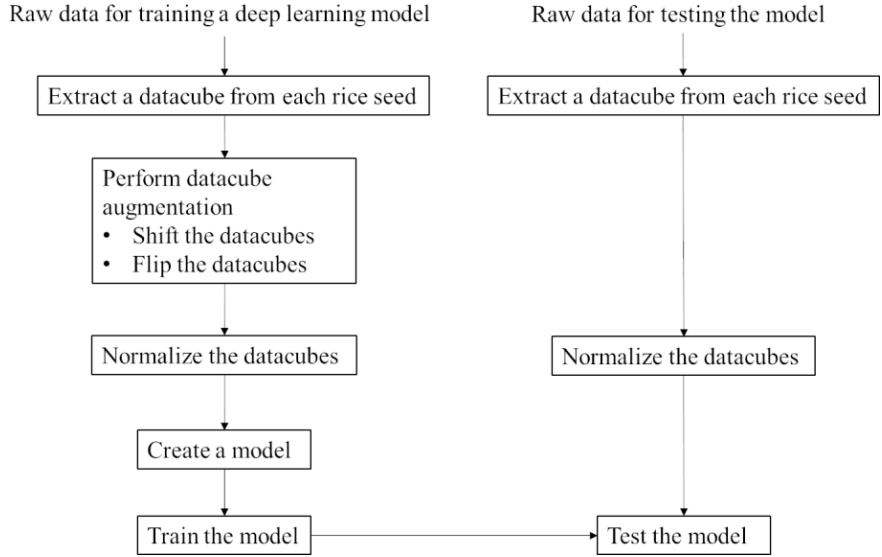

Figure 5. A simplified flowchart of the proposed method.

| Methods | Number of parameters (M) | Number of convolutional layers |
|---------|--------------------------|-------------------------------|
| VGGNet | 35.80 | 31 |
| ResNet | 35.97 | 34 |
| ResNet-B | 35.76 | 116 |

Table 1. Number of parameters and convolutional layers of VGGNet, ResNet, and ResNet-B.

## 3. Results and Discussion

### 3.1 Main Experiments

Tables 2 and 3 list the mean top-1 accuracies, top-2 accuracies, macro-averaged precisions, macro-averaged recalls, and macro-averaged f-scores of the four classification methods on the processed and paddy rice datasets, respectively. Table 4 lists the training and testing times of the four methods under consideration. Figures 6 and 7 show representative confusion matrices for the processed rice and paddy rice datasets, respectively. All of the JP rice seeds were correctly identified even when only the spatial information was used due to their distinct visual appearances. All of the KN rice seeds were perfectly identified using their unique spectral profiles. However, misclassifications were observed for the rice varieties that were more similar spatially or spectrally (such as the LP and ML varieties in the processed rice dataset, and all the varieties in the paddy rice dataset). The misclassification errors were reduced by using both spatial and spectral information. Exploiting spatio-spectral information automatically extracted from the datacubes, ResNet-B achieved the best performances, as measured by the five evaluation metrics, on both datasets.

Since the SVM-based methods need to perform a cascade of data preprocessing steps for feature extractions as described in the Appendix, they could encounter potential error propagation between the steps, which is extremely unlikely for the deep learning-based methods, which do not require these preprocessing steps. For instance, if a far-from-optimal value is used for the adaptive thresholding step, we could get a binary mask that does not represent the rice seed well. Consequently, both spatial and spectral feature extraction processes, which heavily rely on the mask, will be inaccurate, leading to decreased classification accuracy. Similar to other deep learning-based rice classification methods (Lin et al. 2018; Patel and Joshi 2018; Qiu et al. 2018), the proposed method does not rely on hand-engineered features, which could be suboptimal for the given rice variety classification task. For instance, while the hand-engineered spatial features were successfully used to distinguish some of the rice varieties in the processed rice dataset (e.g., the JP variety versus the rest), they were not very useful for the paddy rice dataset because the spatial features of different rice varieties were very similar to each other. This observation suggests that using the predefined features is not the most appropriate approach. Treating feature extraction as a separate process from classification could also lead to loss of information that cannot be recovered since the datacube is typically discarded after the feature extraction process.

| Methods | Top-1 accuracy (%) | Top-2 accuracy (%) | Precision | Recall | F-score |
|---|---|---|---|---|---|
| SVM (spatial) | 73.25 | 92.86 | 0.7304 | 0.7325 | 0.7287 |
| SVM (spectral) | 94.29 | 99.13 | 0.9434 | 0.9429 | 0.9422 |
| SVM (spatial and spectral) | 95.00 | **99.84** | 0.9512 | 0.9492 | 0.9486 |
| ResNet-B | **97.54** | **99.84** | **0.9767** | **0.9754** | **0.9754** |

Table 2. The mean top-1 accuracies, top-2 accuracies, macro-averaged precisions, macro-averaged recalls, and macro-averaged f-scores of the four classification methods on the processed rice dataset (6 repetitions).

| Methods | Top-1 accuracy (%) | Top-2 accuracy (%) | Precision | Recall | F-score |
|---|---|---|---|---|---|
| SVM (spatial) | 50.89 | 75.77 | 0.5108 | 0.5089 | 0.5036 |
| SVM (spectral) | 74.11 | 92.98 | 0.7402 | 0.7411 | 0.7392 |
| SVM (spatial and spectral) | 79.23 | 94.64 | 0.7931 | 0.7923 | 0.7911 |
| ResNet-B | **91.09** | **97.58** | **0.9149** | **0.9109** | **0.9105** |

Table 3. The mean top-1 accuracies, top-2 accuracies, macro-averaged precisions, macro-averaged recalls, and macro-averaged f-scores of the four classification methods on the paddy rice dataset (10 repetitions).

| Methods | Training time | Testing time per batch size of 230 datacubes |
|---|---|---|
| SVM (spatial) | 7.6 s | 0.02 s |
| SVM (spectral) | 18.7 s | 0.05 s |
| SVM (spatial and spectral) | 8.7 s | 0.05 s |
| ResNet-B | 15 hrs | 14.3 s |

Table 4. The training times and testing times averaged over the two datasets.

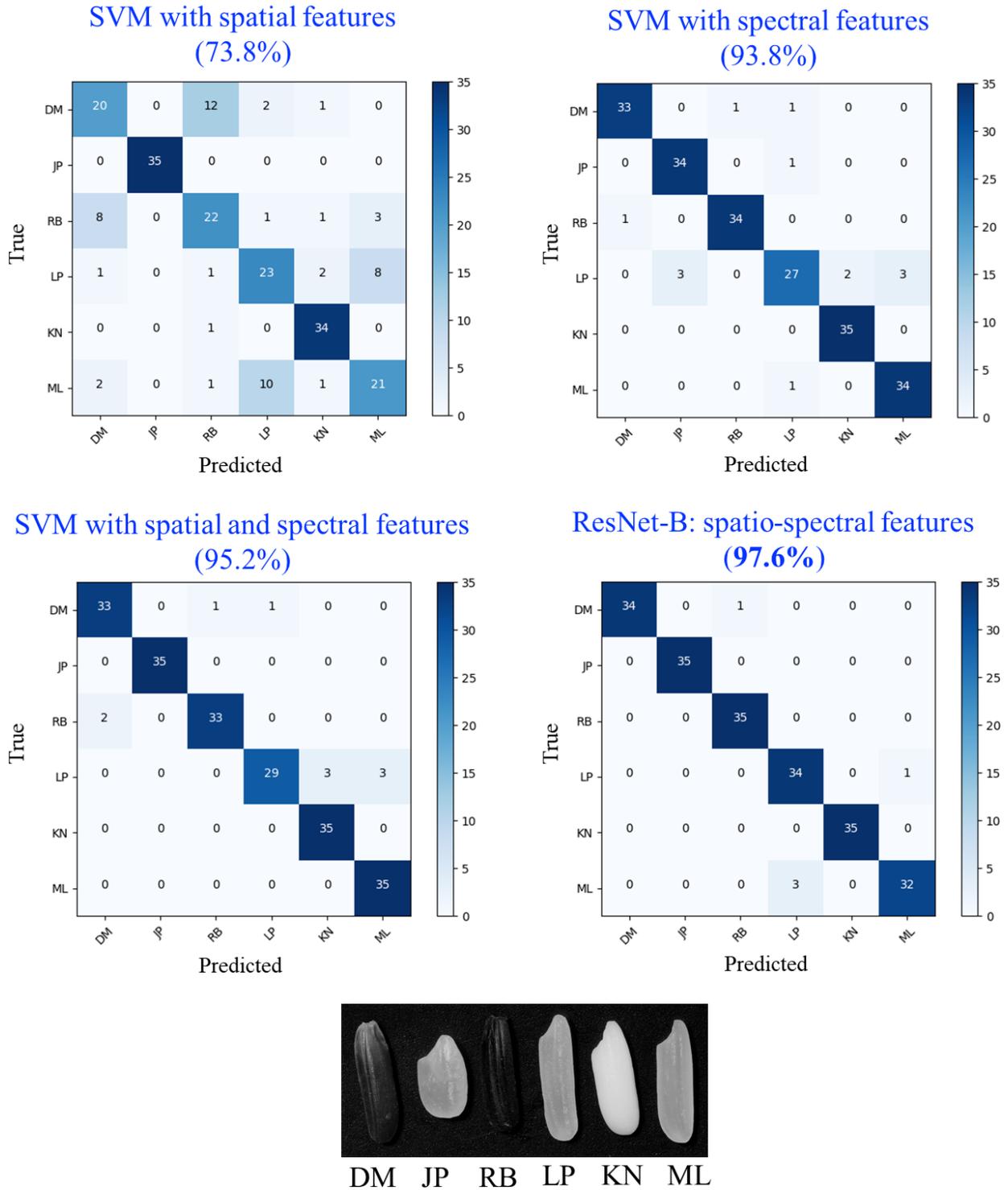

Figure 6. Representative confusion matrices and classification accuracies of the processed rice dataset. The number in the intersection of the $i^{th}$ row and $j^{th}$ column of each matrix is equal to the number of rice seeds of variety $i$ being classified as variety $j$.

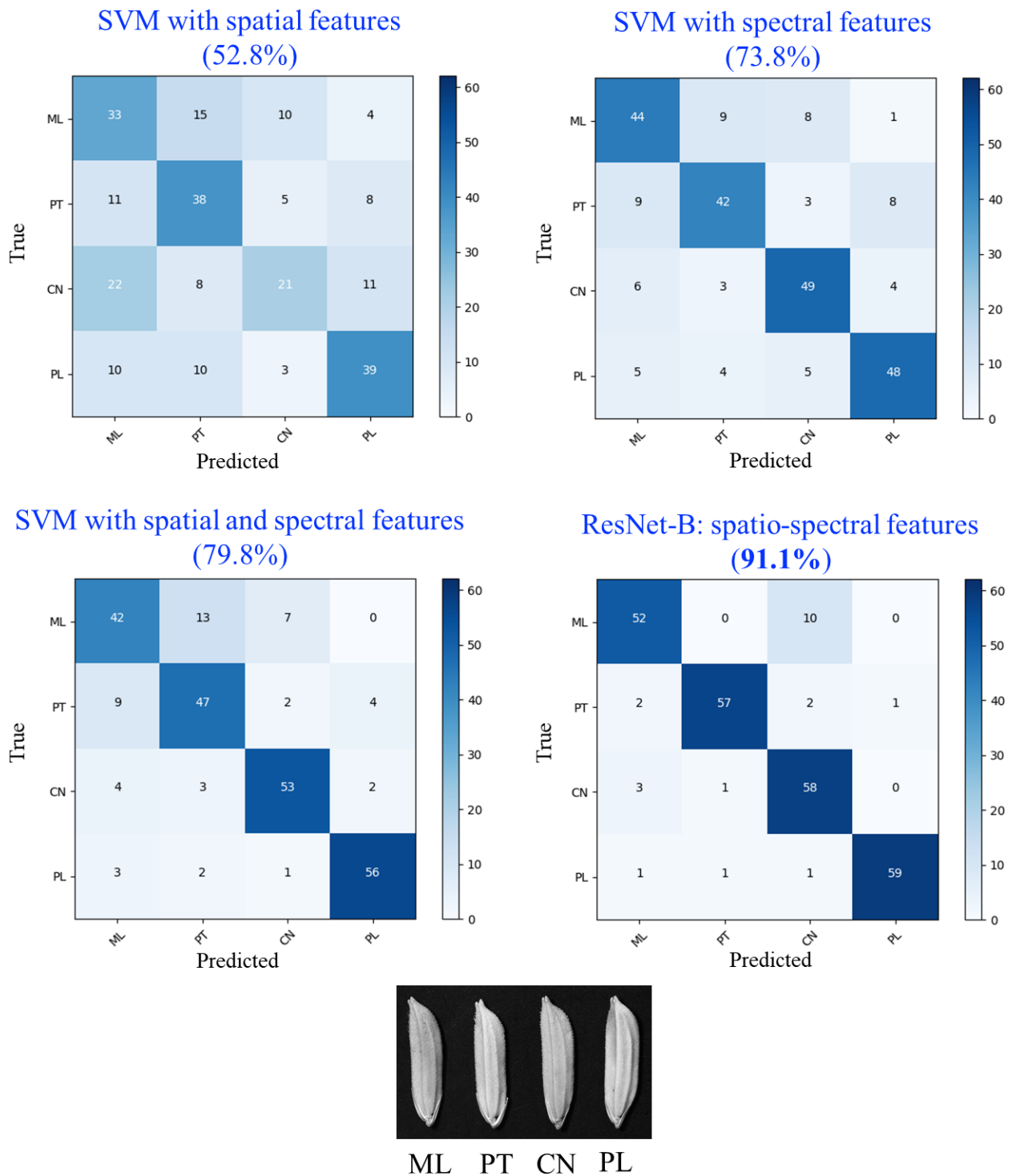

Figure 7. Representative confusion matrices and classification accuracies of the paddy rice dataset. The number in the intersection of the $i^{th}$ row and $j^{th}$ column of each matrix is equal to the number of rice seeds of variety $i$ being classified as variety $j$.

## 3.2 Deep Learning Experiments

Table 5 lists the mean top-1 accuracies, top-2 accuracies, macro-averaged precisions, macro-averaged recalls, and macro-averaged f-scores of the four classification methods, which were developed based on spatio-spectral deep CNN, on the paddy rice dataset. Table 6 lists the training and testing times of the deep learning-based methods. VGGNet had the worst performance, as measured by the five evaluation metrics. Exploiting residual connections, ResNet achieved better performance than VGGNet. Using both residual connections and bottleneck building blocks, ResNet-B further improved the rice classification performance because it contained many more processing layers, leading to higher representation power. The ensemble method, which combined the predictions of VGGNet, ResNet, and ResNet-B, achieved the best performance on all the evaluation metrics. Figure 8 shows the box plots of top-1 classification accuracies obtained from all the methods implemented in this work for the paddy rice dataset. The standard deviations of the top-1 accuracies over 10 repetitions were 3.43 for SVM with spatial features, 2.26 for SVM with spectral features, 1.55 for SVM with spatial and spectral features, 2.09 for VGGNet, 2.66 for ResNet, 2.35 for ResNet-B, and 1.92 for the ensemble method. Among the deep learning based methods, the ensemble method achieved the lowest standard deviation of the top-1 classification accuracy. We have also observed similar trends for all the remaining evaluation metrics.

As shown in Figure 8, the deep learning-based methods had higher variances than some of the SVM-based methods due to the stochastic nature of deep learning models. Different initial values of the weights and biases in deep learning models result in different final models. The training algorithm for deep learning is also stochastic in nature, contributing to increased variances. One effective way to reduce variances in deep learning models is to combine several trained deep learning models into an ensemble (Krizhevsky et al. 2012; Goodfellow et al. 2016). As demonstrated in our experiment, the variances of the deep learning models got reduced by using the ensemble method. It is also worth pointing out that while the variances of the deep learning methods were slightly higher than those of the SVM-based methods, the mean accuracies of the deep learning methods were much higher, leading to consistently higher effective classification accuracies.

| Methods | Top-1 accuracy (%) | Top-2 accuracy (%) | Precision | Recall | F-score |
|---|---|---|---|---|---|
| VGGNet | 86.41 | 96.09 | 0.8702 | 0.8641 | 0.8645 |
| ResNet | 88.23 | 96.21 | 0.8904 | 0.8823 | 0.8823 |
| ResNet-B | 91.09 | 97.58 | 0.9149 | 0.9109 | 0.9105 |
| Ensemble | **93.27** | **97.90** | **0.9339** | **0.9327** | **0.9326** |

Table 5. The mean top-1 accuracies, top-2 accuracies, macro-averaged precisions, macro-averaged recalls, and macro-averaged f-scores of the four classification methods, which were developed based on spatio-spectral deep CNN, on the paddy rice dataset (10 repetitions).

| Methods | Training time | Testing time per batch size of 248 datacubes |
|---------|---------------|----------------------------------------------|
| VGGNet | 9.8 hrs | 6.1 s |
| ResNet | 10.1 hrs | 6.6 s |
| ResNet-B | 16.4 hrs | 15.4 s |

Table 6. The training times and testing times on the paddy rice dataset

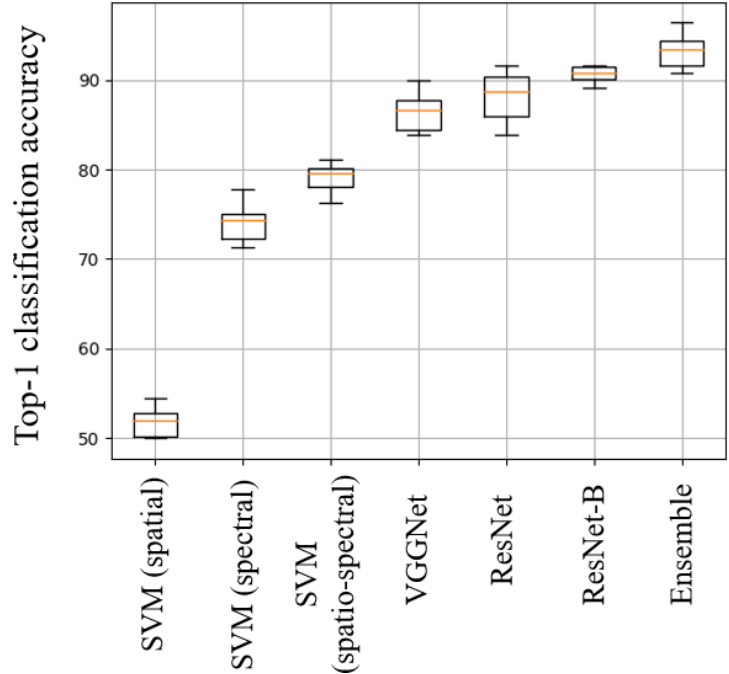

Figure 8. Box plots of top-1 classification accuracies obtained from all the methods under consideration for the paddy rice dataset (10 repetitions).

## 3.3 Extensions and Limitations

Having an ability to learn data representations with multiple levels of abstractions directly from the datacube by itself, the proposed deep learning-based methods are much less susceptible to biases from human past experiences, which might be irrelevant in some cases, and hence could discover data representations that truly matter for classification. However, interpreting a trained deep CNN is complicated. Several methods based on visualizations have been proposed to help us understand a trained network better (Simonyan et al. 2013; Springenberg et al. 2014; Zeiler and Fergus 2014; Yosinski et al. 2015; Selvaraju et al. 2017). In this work, saliency maps were used to gain some insight into how the trained deep CNNs could have made the decision (Figure 9). An image-specific saliency map provides some information on how the decision made by a trained CNN changes with respect to a small change in each input region (Simonyan et al. 2013). The regions (pixels) in the saliency map with high values contribute more towards the decision (i.e., predicted rice variety), compared to the low values regions. As shown in Figure 9, the trained deep CNNs were able to automatically locate the rice seeds in the datacubes and neglect the background regions, bypassing the predefined parameter-dependent rice seed extraction step required in the existing traditional rice variety classification methods (Liu et al. 2005, 2010; Guzman and Peralta 2008; OuYang et al. 2010; Mousavi Rad et al. 2012; Kong et al. 2013; Pazoki et al. 2014; Hong et al. 2015; Sun et al. 2015; Wang et al. 2015; Vu et al. 2016; Kuo et al. 2016; Huang and Chien 2017). Nevertheless, it is not straightforward to tell exactly which specific spatial and/or spectral locations (i.e., voxels) were exploited to distinguish different rice varieties effectively. In general, even with the emerging visualization techniques, it is still challenging to fully make sense of the learned data representations and how the trained network makes a decision, especially for the data with more than two dimensions such as hyperspectral data. Developing new tools to crack open the "black boxes" of deep learning is still an active area of research.

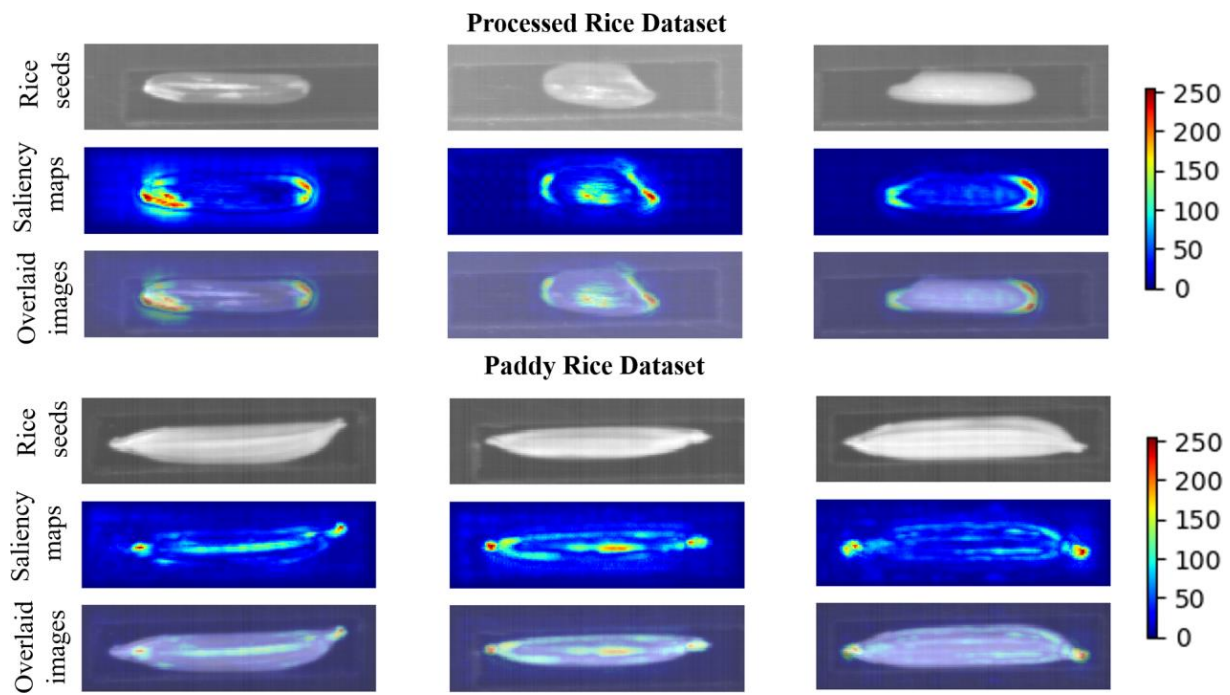

Figure 9. The image-specific saliency maps of some representative rice seeds. The regions with high values in each saliency map contribute more towards the rice variety prediction.

In this work, all the rice seeds were scanned using the hyperspectral imaging system in the same orientation. This fixed-orientation setup simplified the training process of the proposed method. However, for the case when it is more convenient to be able to scan rice seeds in any orientations, it becomes more difficult to train a deep CNN. Specifically, there will be more variations in the datacubes. Consequently, more data will possibly be needed to maintain comparable classification accuracy. In addition to more data acquisition, rotation operations can be included in the data augmentation process to generate more training data.

Although the near-infrared region was selected for this work because of its ability to provide information on chemical properties, which can help distinguish rice varieties, a reasonable extension to the proposed method could be to also acquire information in other regions of the electromagnetic spectrum, and jointly use the complementary information to improve the classification accuracy. Moreover, with the current setup, the proposed system will certainly benefit from higher spatial resolution in terms of classification accuracy, which could be achieved by incorporating a microscope to the data acquisition system along with a speed controller that can reliably maintain the velocity of the conveyor belt, especially when the velocity becomes very low. In this case, there will be a tradeoff between the classification accuracy and throughput of the inspection system.

In addition to the modifications on the data acquisition part, the data processing and classification part could be improved. Since hyperspectral data can be thought of as a series of images over the spectral axis, Recurrent Neural Networks (RNN) with the Long Short Term Memory (LSTM) unit (Hochreiter and Urgen Schmidhuber 1997; Graves and Schmidhuber 2005), Gated Recurrent Unit (GRU) (Cho et al. 2014), and attention mechanism (Bahdanau et al. 2014) would be potential candidates for improved classification performance because these methods are good at dealing with sequential data. While we have demonstrated the potential of several deep learning-based methods and their ensemble for the rice variety classification task, designing and determining an optimal model for this task is beyond the scope of this work and should be investigated in future research.

## 4. Conclusions

In this work, we proposed a non-destructive rice variety classification method that benefited from the synergy between hyperspectral imaging and spatio-spectral deep learning. Hyperspectral imaging provided not only spatial information, but also spectral information of the rice seeds under inspection. A deep CNN that automatically performed spatio-spectral feature extractions without any predefined data processing steps was used to classify the rice variety of each rice seed. As demonstrated using two types of rice datasets, the proposed method (ResNet-B) yielded higher classification accuracies than the most commonly used classification methods based on SVM. The proposed method achieved 91.09% mean classification accuracy for the paddy rice dataset, which consisted of the rice varieties that have been frequently mistaken for the others, compared to 79.23% obtained from SVM with both spatial and spectral information.

## Acknowledgments

We would like to acknowledge the Plant Geoinformatics and Digital Management System Laboratory, National Center for Genetic Engineering and Biotechnology for providing us with the paddy rice seeds. We would like to acknowledge Dr. Thiparat Chotibut for the discussion on the ESP activation function.

# Appendix

## Support Vector Machines (SVM)

SVM is a supervised learning method that is widely used for regression and classification frameworks due to its ability to effectively handle high dimensional data, even when the number of samples is lower than the number of input dimensions (Boser et al. 1992; Cortes and Vapnik 1995). Some information that characterizes the raw data (i.e., features) is typically extracted and used as input for the SVM. If used for classification tasks, SVM constructs a hyperplane or a set of hyperplanes that represents the largest separation of data from different classes in the feature space and uses the constructed hyperplanes to classify unseen test data. The classification accuracy of SVM highly depends on the quality of features extracted from the data. For rice classification, two types of features are typically extracted from each rice seed (Figure A1): spatial features (Liu et al. 2005; Guzman and Peralta 2008; OuYang et al. 2010; Mousavi Rad et al. 2012; Pazoki et al. 2014; Hong et al. 2015; Wang et al. 2015; Sun et al. 2015; Kuo et al. 2016; Vu et al. 2016; Huang and Chien 2017) and spectral features (Liu et al. 2010; Kong et al. 2013; Sun et al. 2015; Wang et al. 2015; Vu et al. 2016). Then, the features are typically standardized and used as SVM inputs to determine the variety of each rice seed.

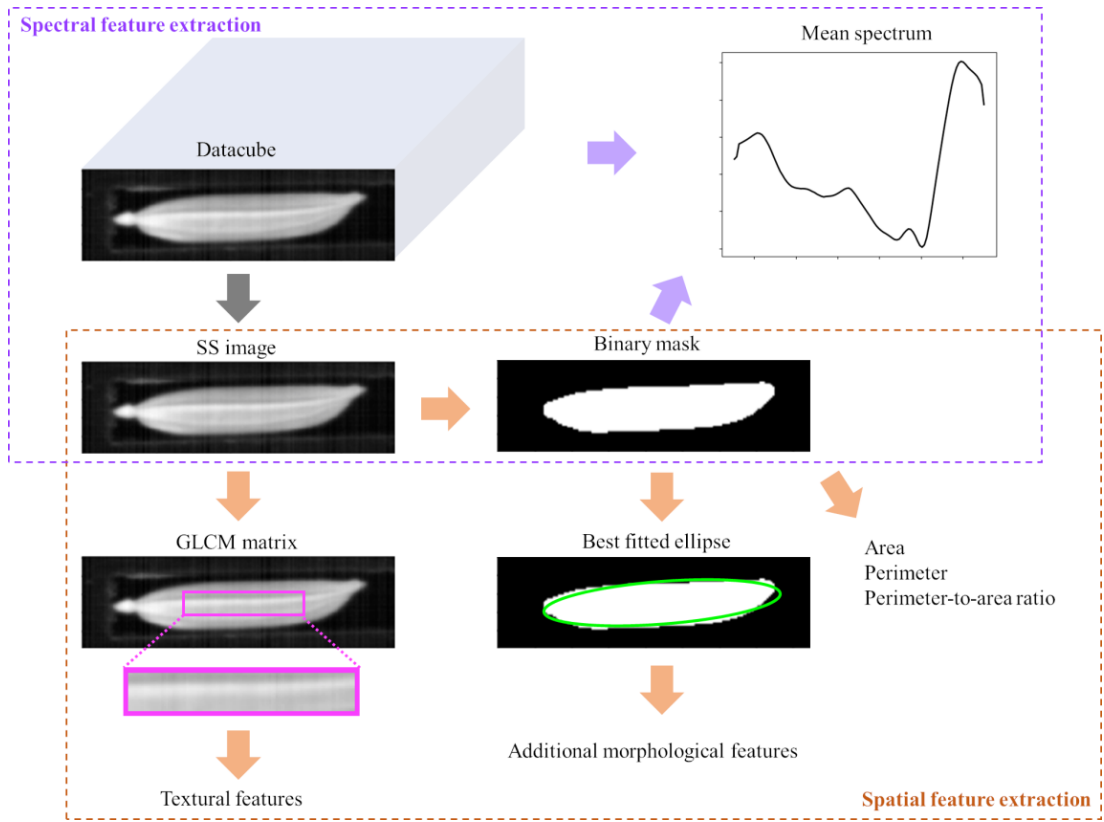

Figure A1. The spatial and spectral feature extraction process for the SVM-based rice classification methods in this work. Multiple sequential data processing steps are performed to extract the features.

## Spatial Feature Extraction

Similar to the spatial features used in recent works (Liu et al. 2005; Guzman and Peralta 2008; OuYang et al. 2010; Mousavi Rad et al. 2012; Pazoki et al. 2014; Hong et al. 2015; Wang et al. 2015; Sun et al. 2015; Kuo et al. 2016; Vu et al. 2016; Huang and Chien 2017), six textural and eleven morphological features were used in this work. For each rice seed, a sum-of-squares (SS) image was computed by summing the squared-magnitudes of the datacube along the spectral dimension.

The textural features were extracted directly from the resulting SS image by computing the following features of a gray-level co-occurrence matrix (Haralick et al. 1973) as defined in (Van der Walt et al. 2014): contrast, dissimilarity, homogeneity, angular second moment, energy, and correlation.

For morphological features, three data processing methods were sequentially performed on the SS image to get a clean binary image (also called a mask) that will be used for morphological feature extractions. First, the SS image was filtered using a Gaussian kernel to mitigate unwanted artifacts. Then, adaptive thresholding was performed to get an initial mask (1 for the pixels that belong to the rice seed, and 0 otherwise). Finally, morphological opening and closing operations were performed to clean up the mask.

With the resulting mask, eleven morphological features were extracted: surface area, perimeter, perimeter-to-area ratio, major axis length, minor axis length, first eccentricity, standard deviation of the radii, minimum radius, maximum radius, maximum-to-minimum radius ratio, and Haralick ratio. While the surface area, perimeter, and perimeter-to-area ratio were calculated directly from the mask, an additional processing step was required to extract the other morphological features. Specifically, an ellipse was fit to the mask to approximately represent the rice seed boundary. The ellipse with the lowest fitting error in a least-squares sense was kept. The radii were defined as the distances from the centroid of the ellipse to the contour points. The Haralick ratio was defined as the ratio of the mean radius to the standard deviation of the radii.

## Spectral Feature Extraction

While spatial features provide information on the visual appearances, spectral features provide complementary information on chemical properties of the rice seeds. The mean spectrum of each datacube over a region of interest (ROI) was used as spectral features. In this work, an ROI was chosen to be the region that contained all voxels within the rice seed.

## Implementation Details for the Main Experiments

Three types of features were used as inputs to SVMs: spatial, spectral, and spatio-spectral features. The spatial features and spectral features were extracted from the datacubes as explained above. Following existing works (Sun et al. 2015; Wang et al. 2015; Vu et al. 2016), a simple concatenation of the spatial and spectral features was treated as the spatio-spectral features. For all the three SVM-based classification methods, a kernel trick (Boser et al. 1992) was also used with the radial basis function kernel. The hyperparameters associated with the SVM methods were determined using cross-validation with stratified random splits on the training data with five splitting and reshuffling iterations. After the hyperparameters had been selected, the internal parameters of the classifiers (e.g., weights and biases) were optimized using all of the training data. Finally, the performances of the trained classifiers were assessed using the test dataset.